\documentclass[sigconf,screen]{acmart}

\AtBeginDocument{%
  \providecommand\BibTeX{{%
    \normalfont B\kern-0.5em{\scshape i\kern-0.25em b}\kern-0.8em\TeX}}}

\setcopyright{none}
\acmConference[Preprint]{Preprint}{2021}{}

\usepackage{soul}
\usepackage{mathrsfs}
\usepackage{diagbox}
\usepackage{multirow}
\graphicspath{{./figure/}}

\usepackage{enumerate}
\usepackage[ruled,linesnumbered]{algorithm2e}
\usepackage{subcaption}

\author{Xin Guo}
\affiliation{%
  \department{School of Computer Science and Engineering, BDBC}
  \institution{Beihang University}
  \city{Beijing}
  \country{China}
}
\email{guoxin@act.buaa.edu.cn}

\author{Jianlei Yang}
\affiliation{%
  \department{School of Computer Science and Engineering, BDBC}
  \institution{Beihang University}
  \city{Beijing}
  \country{China}
}
\email{jianlei@buaa.edu.cn}
\author{Haoyi Zhou}
\affiliation{%
  \department{School of Computer Science and Engineering, BDBC}
  \institution{Beihang University}
  \city{Beijing}
  \country{China}
}
\email{zhouhy@act.buaa.edu.cn}
\author{Xucheng Ye}
\affiliation{%
  \department{School of Computer Science and Engineering, BDBC}
  \institution{Beihang University}
  \city{Beijing}
  \country{China}
}
\email{zy1806121@buaa.edu.cn}
\author{Jianxin Li}
\affiliation{%
  \department{School of Computer Science and Engineering, BDBC}
  \institution{Beihang University}
  \city{Beijing}
  \country{China}
}
\email{lijx@act.buaa.edu.cn}

\begin{document}

\title{RoSearch: Search for Robust Student Architectures When Distilling Pre-trained Language Models}

\renewcommand{\shortauthors}{Xin and Jianlei, et al.}

\begin{abstract}

Pre-trained language models achieve outstanding performance in NLP tasks. Various knowledge distillation methods have been proposed to reduce the heavy computation and storage requirements of pre-trained language models. However, from our observations, student models acquired by knowledge distillation suffer from adversarial attacks, which limits their usage in security sensitive scenarios. In order to overcome these security problems, \texttt{RoSearch} is proposed as a comprehensive framework to search the student models with better adversarial robustness when performing knowledge distillation. A directed acyclic graph based search space is built and an evolutionary search strategy is utilized to guide the searching approach. Each searched architecture is trained by knowledge distillation on pre-trained language model and then evaluated under a robustness-, accuracy- and efficiency-aware metric as environmental fitness. Experimental results show that \texttt{RoSearch} can improve robustness of student models from $7\%$\textasciitilde$18\%$ up to $45.8\%$\textasciitilde$47.8\%$ on different datasets with comparable weight compression ratio to existing distillation methods ($4.6\times$\textasciitilde$6.5\times$ improvement from teacher model $\rm BERT_{BASE}$) and low accuracy drop. In addition, we summarize the relationship between student architecture and robustness through statistics of searched models.




\end{abstract}



\maketitle

\section{Introduction}
\label{sec:intro}

Recently pre-trained language models have become as the mainstreaming approaches in Natural Language Processing (NLP) tasks, such as BERT~\cite{bert}, XLNet~\cite{xlnet}, RoBERTa~\cite{roberta}, etc.
By being pre-trained on large scale corpus datasets, these models extract very rich knowledge-abilities from natural language texts.
And then some fine-tuning processes are usually performed with task-specific labeled data to provide the knowledge-abilities for solving downstream NLP tasks.
The combination of natural language knowledge and task-aware knowledge makes it possible for pre-trained language models to refresh state-of-the-art performance in many NLP tasks.

With the growth in demand for performance on NLP tasks, the scale of pre-trained models increase in terms of both parameter number and inference latency.
The increasing scale of pre-trained models bring very high cost both in computation and memory for practical usage.
Especially for the mobile or embedded devices, it is difficult to deploy pre-trained models because of the very limited hardware resources.
A popular approach is to make those powerful models lightweight for easier deployments on edge devices.
Many model compression techniques have been proposed~\cite{albert,comp_bert,distill1,x1,x2,x4,x5,x6} to reduce their model size while maintaining model accuracy.
Among those model compression approaches, knowledge distillation attracts much attention and becomes a mainstream research direction.
Knowledge distillation was proposed in~\cite{hinton} as a teacher-student framework, in which student models are trained to imitate the behavior of teacher models.
Knowledge distillation could improve model's efficiency by transferring its knowledge to lightweight neural architectures.
Compared to being trained from scratch, student models usually perform better when trained with knowledge distillation methods.
To fully utilize the rich knowledge-abilities in pre-trained language models, various knowledge distillation methods have been proposed, especially for the well-known pre-trained model BERT~\cite{bert}.


\begin{figure}[t]
\begin{center}
\vspace{5mm}
\centerline{\includegraphics[width=\columnwidth]{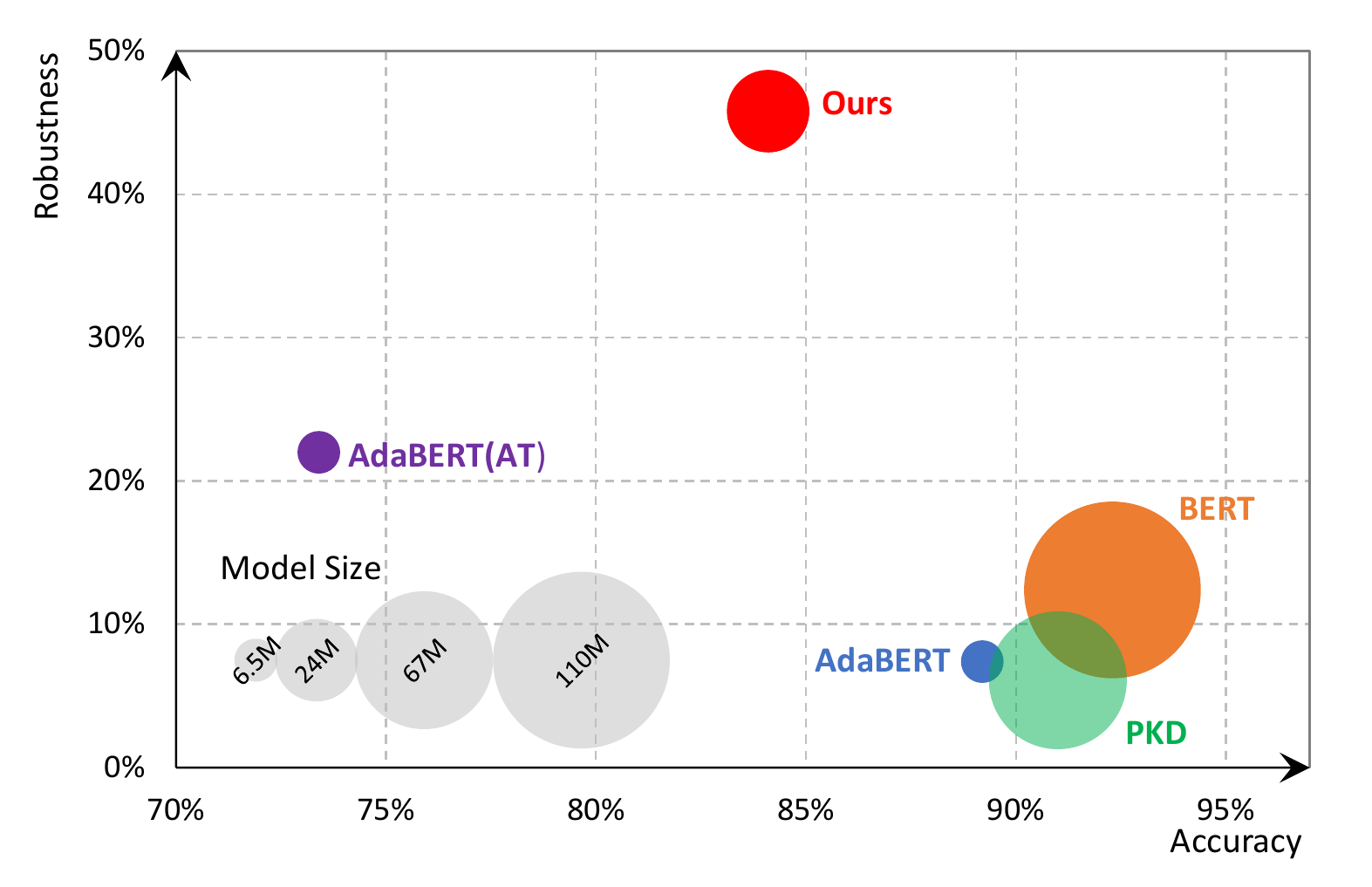}}
\vspace{-2mm}
\caption{Robustness, accuracy and model size comparison of pre-trained model BERT~\cite{bert}, distilled model PKD~\cite{pkd}, AdaBERT~\cite{adabert}, AdaBERT with adversarial training (AT) and our method on SST-2 dataset. Robustness is measured by TextFooler~\cite{textfooler} attack method. \protect\footnotemark}
\vspace{-5mm}
\label{fig:bubble}
\end{center}
\end{figure}

\footnotetext{AdaBERT-SST2 is re-implementation of knowledge distillation with fixed architecture provided in~\cite{adabert}.}

Existing distillation efforts mainly focus on how to improve the model's efficiency while maintaining the accuracy.
The models can be compressed up to $15\times$ with negligible accuracy drop~\cite{adabert}.
However, another important issue is raised: are these student models with good accuracy and efficiency suitable enough for real-world applications?
It has been explored that neural networks are usually vulnerable to adversarial attacks~\cite{aa01,aa02}. 
Minimal perturbation on input sample may lead to wrong classification. 
For pre-trained language models, TextFooler demonstrated the possibility to generate adversarial examples for the well-known BERT~\cite{textfooler}. 
An intuitive concern arises whether students distilled from pre-trained models suffer from adversarial attacks.
Several knowledge distillation works on pre-trained models have been evaluated for comparing their robustness and accuracy as shown in Figure~\ref{fig:bubble}.
Lightweighted or distilled models could have a good accuracy (\textasciitilde$90\%$), but all of them pose a very low robustness ($6\%$\textasciitilde$7\%$) under adversarial attacks.
Such a low robustness implies a strong vulnerability of student models, which will bring potential risk to their practical usage in security sensitive scenarios. 



In real-world applications, efficiency, accuracy and robustness are all crucial metrics for neural network models.
Pre-trained language models perform well in terms of accuracy while knowledge distillation techniques improve their efficiency, but robustness is left behind in existing researches.
In this work, we aim to reduce the potential security risk of these student models and perform appropriate balancing among accuracy, efficiency and robustness.
Since improving the robustness of large pre-trained models might be very resource consuming, and it is difficult to transfer teacher's robustness to students (as experiments demonstrated in Section \ref{subsec:modeleval}), we mainly pay attention to improving the robustness when performing knowledge distillation.


\begin{figure}[t]
\begin{center}
\centerline{\includegraphics[width=\columnwidth]{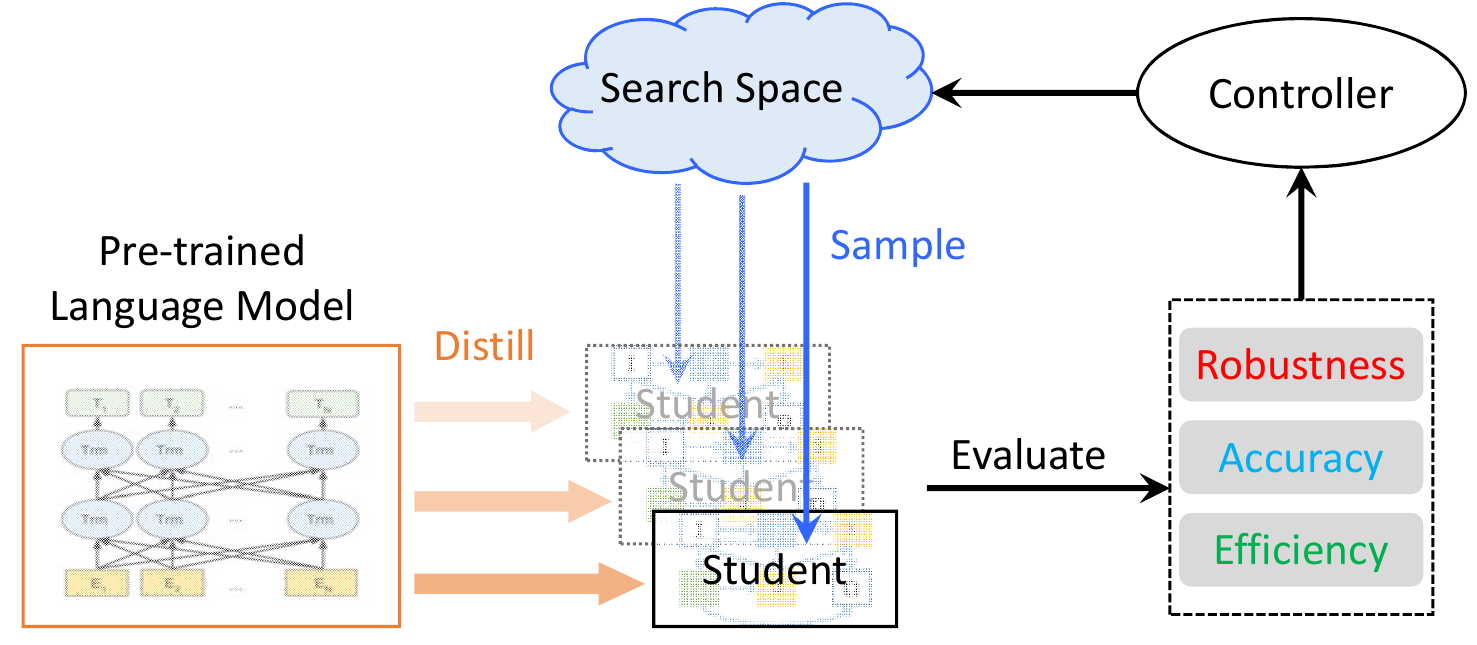}}
\caption{Overview of \texttt{RoSearch}. Student architectures are sampled from the search space. Pre-trained model is performed by knowledge distillation on these samples. And then these students architectures are evaluated under a robustness-, accuracy- and efficiency-aware metric. NAS controller determines the searching actions according to the evaluation feedbacks on metric.}
\vspace{-8mm}
\label{fig:overview}
\end{center}
\end{figure}

Our research mainly focuses on exploring student models in neural architecture view.
Knowledge distillation could improve the performance of student models in their weights optimizing stage, but their neural architectures are not well studied.
The neural architectures of student models are usually inherited from large-scale pre-trained models, like BERT~\cite{bert}.
There lies some attempts~\cite{adabert} on applying neural architecture search (NAS) approach in pre-trained model distillation, their search objective is elaborate designed on improving performance and efficiency rather than enhancing the model robustness.
In this work, an evolutionary NAS algorithm is exploited for student architectures search with considering high robustness. And the relationship between student architectures and robustness is also explored.
This paper aims to answer these Questions:
\begin{enumerate}
\item[\large \textcircled{\small \textbf{1}}] How about the robustness characteristics of student models after knowledge distillation?
\item[\large \textcircled{\small \textbf{2}}] How to improve the robustness of student models?
\item[\large \textcircled{\small \textbf{3}}] How about the relationship between robustness and searched student architectures?
\end{enumerate}

Aiming to improve the robustness of student models, \texttt{RoSearch} is proposed as a NAS framework as shown in Figure~\ref{fig:overview}.
The pre-trained language model is distilled as student models, while student architecture is searched under a robustness-, accuracy- and efficiency-aware model evaluation metric.
An evolutionary search strategy is utilized to guide the searching procedures according to the evaluation feedbacks, similar to ~\cite{evo_trans}.
A block-based search space is built by each block represented by a Directed Acyclic Graph (DAG) whose vertices and edges represent the network layers and data-flow respectively.
To guide the searching towards a more robustness direction, we generate adversarial examples for each model and calculate the accuracy of distilled student model as robustness metric.
The metrics of accuracy on benign datasets and efficiency measured by parameter number are also included to maintain the advantage of pre-trained language models and knowledge distillation.

The key contributions of this paper are summarized as follows:
\begin{itemize}
\item We demonstrate that knowledge distillation methods for pre-trained language models suffer from potential security risk brought by adversarial attacks.
\item We propose a \texttt{RoSearch} framework that can improve robustness of student models while maintaining accuracy and efficiency characteristic in knowledge distillation setting.
\item We summarize the relationship between robustness and architectures of searched student models, which could help guiding the robust search on pre-trained models.
\end{itemize}

The rest of this paper is organized as follows. Section \ref{sec:related} reviews the related works. Problem formulations are demonstrated in Section \ref{sec:formulation}. Section \ref{sec:notrobust} indicates existing robustness issues of distillation. Details of \texttt{RoSearch} are provided in Section \ref{sec:archsearch}. Section \ref{sec:experiments} illustrates experimental results. Concluding remarks are given in Section \ref{sec:conclusions}.

\section{Related Works}
\label{sec:related}

\textbf{Pre-trained Language Models Distillation.}
Research of knowledge distillation of pre-trained language models starts from~\cite{distill1}.
They proposed an approach to distill knowledge from BERT to BiLSTM student model.
Mean Squared Error Loss between logits and Cross Entropy Loss on true labels are adopted in distillation.
Similar to the architecture of BERT, stacking Transformer encoder layers~\cite{transformer} are utilized as student models, such as~\cite{distilbert,pkd,tinybert}.
And NAS approach is utilized to find better CNN-based student architectures for specific downstream tasks~\cite{adabert}.
To the best of our knowledge, none of such related works has taken robustness into account.

\textbf{Robustness-aware NAS.}
RobNet is proposed as a robustness-aware search algorithm based on one-shot NAS approach~\cite{whennasmeets}, which evaluates the robustness of each model after vanilla training and adversarial fine-tuning. 
A robustness indicator was designed to accelerate searching by discarding the architectures with lower possible robustness. 
Relationship between architecture patterns and adversarial robustness in image classification tasks is also discussed.
\cite{evolvingrobust} proposed another robustness-aware NAS.
Layer-, block- and model-level evolution algorithm was combined to fully explore robustness in different level of architectures.
Transferability based black-box robustness evaluation was applied on each architecture while searching.
Adversarial examples were generated before searching to reduce total runtime.
However, these existing works focused on computer vision area, convolutional network architectures and vanilla training without knowledge distillation. 

\textbf{Adversarial Training.}
Adversarial training is a common approach to improve robustness of neural networks.
FGSM~\cite{aa02} algorithm is proposed to generate adversarial examples at low computational cost at training stage.
Generated adversarial examples are appended into training set, so that neural networks can adapt to these malicious data and acquire higher robustness after training, that is, adversarial training.
\cite{madry} utilizes a stronger attack algorithm, projected gradient descent method, instead of FGSM to improve the effectiveness of adversarial training.
Adversarial training methods usually work well when training neural networks from scratch.
In this work, the adversarial training is adopted in task fine-tuning stage of pre-trained model and after distillation stage for student models as comparison baseline.

\textbf{Robust Distillation.} 
\cite{ard} evaluated the transferability of robustness from teacher model to student model in knowledge distillation on some image classification datasets. 
Experiments show that the transferability differs according to teacher models and datasets. 
They proposed ARD method to improve the transferability of robustness by punishing distance between student's prediction on adversarial examples and teacher's prediction on benign samples. 
Our work does not focus on distillation loss but the architecture of student models, which is orthogonal to their approach.

\section{Problem Formulation}
\label{sec:formulation}

\textbf{Knowledge Distillation} methods transfer knowledge from teacher model to student model by adding loss terms that force student model to imitate teacher models' behavior in training process.
Traditional distillation computes KL-divergence between student's and teacher's prediction probability distribution softened by temperature $t$:
\begin{gather}
\mathcal{L}_{TKD}=t^2\cdot KL\left(softmax\left(\frac{\mathbf{z}_S}{t}\right),softmax\left(\frac{\mathbf{z}_T}{t}\right)\right),
\end{gather}
where $\mathcal{L}_{TKD}$ is traditional knowledge distillation loss, $\mathbf{z}_S$ and $\mathbf{z}_T$ denote logits output of teacher and student models respectively.

Recent knowledge distillation researches for pre-trained models use additional Cross Entropy Loss, Mean Squared Error (MSE) Loss or other loss functions between student and teacher models' logits, classification outputs, hidden states, attention matrices, or word embeddings.
These loss terms promote knowledge transfer and we summarize all of them by minimizing the weighted distance between teacher and student models as:
\begin{gather*}
\mathcal{L}_{KD}=\lambda_1\mathcal{L}_{logits}+\lambda_2\mathcal{L}_{cls}+\lambda_3\mathcal{L}_{hid}+\lambda_4\mathcal{L}_{attn}+\lambda_5\mathcal{L}_{embd}.
\end{gather*}

\textbf{Model Robustness} measures model's defense ability to adversarial attacks. 
The most commonly used metric of model robustness is to calculate the accuracy of a model on adversarial examples. Adversarial examples are generated from benign inputs to attack target model. 
The more samples in a benign dataset from which can adversarial examples be successfully generated, means the easier a model to be attacked, in other words, the less robust is the model. 
Therefore, robustness of a model $S$ can be defined as:
\begin{gather*}
Robustness(S) \ {\buildrel \Delta \over =} \  \mathcal{A}_S\left(D_{adv}^S\right) ,\ where  \\
D_{adv}^S=\left\{(x',y)\Big| (x,y)\in D, x'=\mathop{\arg\max} \limits_{x' \in \mathcal{N}(x)} \mathcal{L}_{atk}\left(S,x',y\right) \right\},
\end{gather*}
where $S$ denotes student model, $D$ denotes original dataset, $\mathcal{A}_S(\cdot)$ denotes accuracy of student model on a dataset, $D_{adv}^S$ denotes adversarial dataset generated against model $S$.
$\mathcal{N}(\cdot)$ denotes neighbourhood of sample, $\mathcal{L}_{atk}$ denotes loss function that attackers want to maximize for generating adversarial examples.

\textbf{Target of \texttt{RoSearch}}.
As the goal of our research, robustness, accuracy and efficiency are considered as metrics of student models. 
We want to acquire a best student model $S$ with highest score combined by these metrics. 
Accuracy can be easily measured by inference on a dataset $D$. 
Various efficiency metrics are adopted in different researches, such as parameters, MACs, FLOPs, inference time on specific devices, etc. 
We choose the amount of parameters which related to both spacial and temporal complexity as the efficiency metric.
Thus, the target problem can be formulated as:
\begin{gather*}
\label{eq:problem}
\mathop{\arg\max}\limits_{S=(\alpha,\omega^*)} \mathcal{F}\left(\mathcal{A}_S\left(D\right), \mathcal{A}_S\left(D_{adv}^S\right), \#\omega^*\right),  \\
where \  \omega^*=\mathop{\arg\min} \limits_{\omega} \sum \limits_{(x,y)\in D} {\mathcal{L}_{KD}^\alpha(x,y)},
\end{gather*}
where student model $S$ is represented as architecture $\alpha$ and weights $\omega$. 
$\#\omega^*$ denotes number of weights in distilled model and $\mathcal {F}\left(\cdot,\cdot,\cdot\right)$ denotes the function to combine the three metrics.
$\omega$ is updated by minimizing certain knowledge distillation loss function. 
Thus we mainly focus on maximizing the target score by updating $\alpha$, more specifically, by searching student neural architectures.

\section{Conventional Distillation is NOT Robust}
\label{sec:notrobust}

This section aims to answer the Question {\large \textcircled{\small \textbf{1}}} in Section~\ref{sec:intro}.
First, we re-implement several conventional knowledge distillation methods for pre-trained models, and evaluate their robustness under direct attack and transfer attack~\cite{transfer} using TextFooler attack method~\cite{textfooler}. 
Not only the student models distilled by existing methods are applied for directly attack, but also we evaluate their robustness after an adversarial fine-tuning process. 
At the end, we try to figure out whether robustness of teacher model could be transferred to student models.

\subsection{Distillation Is Not Robust}

Existing knowledge distillation methods for pre-trained language models are not robust. 
Several of them are re-implemented  and evaluated on SST-2 dataset. 
For all of those distillations, fine-tuned $\rm BERT_{BASE}$ is adopted as teacher model. 
TextFooler attack method is utilized to evaluate the robustness of student models. 
From the perspective of target model, adversarial attacks can be divided as direct attack and transfer attack. Robustness of each student model is evaluated in both situations. 

Direct attack robustness of each model is measured by its accuracy on adversarial examples generated against the model itself. 
It is the most commonly used metric for model robustness evaluation.
As shown in Table~\ref{tab:distill_direct}, the accuracy of student models distilled from BERT decreased from \textasciitilde$90\%$ on benign dataset to $6\%$\textasciitilde$7\%$ on datasets generated by direct attack. 
Even if compared with teacher model BERT, which is shown not robust, student models suffer more from direct attack. 
The average word changing rate of adversarial examples and average queries required to generate adversarial examples also share the same trend.

\begin{table}[t]
\caption{Robustness of BERT and distilled students under direct attack. $\mathcal{A}$ means the model accuracy. $W\%$ denotes word change rate, $\#Q$ denotes average query times to generate adversarial examples.}
\label{tab:distill_direct}
\vspace{-2mm}
\begin{center}
\begin{sc}
\begin{small}
\begin{tabular}{lcccc}
\toprule
Model	& $\mathcal{A}$ 	& $\mathcal{A}\left(D_{adv}\right)$	& $W\%$	& $\#Q$\\
\midrule
BERT	& $92.3\%$	& $12.4\%$	& $19.3\%$	& $169$ \\
AdaBERT	& $89.0\%$	& $7.1\%$	& $14.0\%$	& $113$ \\
PKD	    & $90.7\%$	& $6.2\%$	& $14.7\%$	& $121$ \\
\bottomrule
\end{tabular}
\end{small}
\end{sc}
\vspace{-2mm}
\end{center}
\end{table}

Intuitively, we believe there are three main reasons of this phenomenon. 
First of all, knowledge distillation aims at compressing the model size of pre-trained language models, in other words, the student models always have fewer weight parameters than original model. 
As a result, the model capacity was also compressed to just fit the natural data in training set.
According to previous research of adversarial robustness in computer vision area, model capacity may influence the model robustness~\cite{madry}. 
Second, student models were not trained from scratch but learned knowledge from teacher model by knowledge distillation approaches. 
Since teacher model was not robust enough, we cannot expect student models reach high robustness.
Accuracy drop, though researches try to avoid, is also common in knowledge distillation results, which may also decrease accuracy under adversarial settings. 
At last, the student model was not designed to be robust and no defense techniques was implemented. 
Traditional distillation researches use artificial designed architecture for student model. 
AdaBERT~\cite{adabert} leverages gradient based search algorithm to search for student architecture, but the target only cares about accuracy on natural data and model efficiency, robustness was not guaranteed since it was not included in evaluation metric.

\begin{table}[t]
\caption{Robustness of BERT and distilled students under transfer attack. Only successful adversarial examples on surrogate model are taken into account.}
\label{tab:distill_transfer}
\vspace{-2mm}
\begin{center}
\begin{sc}
\begin{small}
\begin{tabular}{|l|c|c|c|}
\hline
\diagbox{Surrogate}{$\mathcal{A}\left(D_{adv}\right)$}{Target}	& AdaBERT	& PKD \\
\hline
BERT	&   $53.7\%$	&    $56.4\%$   \\
\hline
AdaBERT	&   $0$	        &    $61.0\%$   \\
\hline
PKD	    &   $56.5\%$	&    $0$        \\
\hline
\end{tabular}
\end{small}
\end{sc}
\vspace{-2mm}
\end{center}
\end{table}

It has been demonstrated that adversarial examples usually have transfer-ability between different models~\cite{transfer}.
Using adversarial examples generated against a known model to attack unknown target model is called transfer attack.
Since knowledge distillation methods and student architectures differ from each other, which increases the difficulty of direct attack, transfer attack from well-known pre-trained models might be a practical risk.
We also evaluate the robustness of student models obtained from existing knowledge distillation methods in the scene that teacher model ($\rm BERT_{BASE}$) or other student models is chosen as surrogate model. 
Results of transfer attack are shown in Table~\ref{tab:distill_transfer}. 
Experiments indicate that, though student models are relatively more robust under transfer attack than direct attack, the accuracy also significantly drops compared with on benign data.

\subsection{Adversarial Training Causes Accuracy Drop}

\begin{table}[t]
\caption{Robustness of BERT and distilled students after adversarial fine-tuning.}
\label{tab:distill_advtrain}
\vspace{-2mm}
\begin{center}
\begin{sc}
\begin{small}
\begin{tabular}{llccc}
\toprule
Model	& Training Method & $\mathcal{A}$ & $\mathcal{A}\left(D_{adv}\right)$ & $W\%$ \\
\midrule
BERT	& Norm. Fine-tune	    & $92.3\%$	& $12.4\%$	& $19.3\%$ \\
BERT	& Adv. Fine-tune	    & $92.1\%$	& $21.3\%$	& $19.8\%$ \\
AdaBERT	& Knowledge Distill	    & $89.0\%$	&  $7.1\%$	& $14.0\%$ \\
AdaBERT	& KD + Adv. Fine-tune	& $73.4\%$	& $22.0\%$	& $16.8\%$ \\
\bottomrule
\end{tabular}
\end{small}
\end{sc}
\vspace{-2mm}
\end{center}
\end{table}

Adversarial training is commonly used to improve robustness of neural networks in their training stage. 
Adversarial examples are generated and appended into training datasets to increase the distance between category border and benign samples. 
In the case of knowledge distillation for pre-trained model, we apply an adversarial fine-tuning stage after distillation. 
Since teacher models are not robust enough, imitating their behaviours cannot provide effective contribution to student model's robustness. 
Thus, only the classification loss on true label is adopted for adversarial fine-tuning stage.
Robustness of pre-trained language model BERT with or without adversarial training tricks in fine-tuning stage is also evaluated for comparison.
As the results shown in Table~\ref{tab:distill_advtrain}, though adversarial training can improve the robustness of BERT while keeping its accuracy, accuracy of student model still drops a lot when robustness improves.

\subsection{Robustness of Teacher Does Not Transfer}

\begin{table}[t]
\caption{Robustness of AdaBERT-SST2 distilled with BERT or adversarial trained BERT as teacher model. Besides only using origin data in knowledge distillation, we also test to append adversarial data to help robustness transfer.}
\label{tab:distill_robust}
\vspace{-2mm}
\begin{center}
\begin{sc}
\begin{small}
\begin{tabular}{|l|c|c|c|c|}
\hline
\multirow{2}{*}{\diagbox{Data}{Teacher}}	& \multicolumn{2}{c|}{BERT} & \multicolumn{2}{c|}{Adv. Trained BERT} \\
\cline{2-5} 
 & $\mathcal{A}$ & $\mathcal{A}\left(D_{adv}\right)$ & $\mathcal{A}$ & $\mathcal{A}\left(D_{adv}\right)$ \\
\hline
Ori. Data    	& $89.0\%$	& $7.1\%$	& $87.5\%$	& $7.3\%$ \\
\hline
+Adv. Data		& $87.3\%$	& $7.6\%$	& $86.5\%$	& $7.9\%$ \\
\hline
\end{tabular}
\end{small}
\end{sc}
\vspace{-2mm}
\end{center}
\end{table}

Though the robustness of pre-trained models (teacher model in distillation) can be improved by adversarial training in fine-tuning stage, we demonstrate that it is difficult to transfer the robustness to student models. 
To experiment whether robustness of pre-trained model can transfer to student models in knowledge distillations, adversarial fine-tuning is performed on BERT and then the adversarial fine-tuned BERT is utilized as teacher model. 
In order to transfer robustness of teacher model to student model as much as possible, the adversarial examples generated in adversarial fine-tuning process is also regarded as a transfer dataset.
However, evaluation results in Table~\ref{tab:distill_robust} show that robustness of adversarial trained teacher model hardly transfer to student model, no matter whether adversarial data are utilized.

\subsection{Summarization}

The robustness characteristics of student models after knowledge distillation are evaluated in this section. 
As a summarization, conventional knowledge distillation methods for pre-trained models are not robust, especially under direct attack.
Even if compared to the teacher model with not a good robustness, lightweight student models still suffer more from adversarial attacks. 
Though adversarial training can improve robustness for student models, it usually causes significant accuracy drop on benign dataset.
Adversarial trained BERT model could obtain robustness improvement while maintaining accuracy, however it is difficult to transfer the robustness to student models.

\section{Architecture Search for Robust Students}
\label{sec:archsearch}

To overcome the robustness problem mentioned in Section~\ref{sec:notrobust}, we propose \texttt{RoSearch}, a comprehensive framework to improve robustness of knowledge distillation by searching student architectures.
Question {\large \textcircled{\small \textbf{2}}} in Section~\ref{sec:intro} will be answered in this section.

\subsection{\texttt{RoSearch} Overview}

As the target problem formulated in Section~\ref{sec:formulation}, we decompose the problem as a two-stage optimization problem: 
\begin{gather*}
\mathop{\arg\max}\limits_{\alpha \in \mathcal{X}} \mathcal{F}\left(\mathcal{A}_S\left(D\right), \mathcal{A}_S\left(D_{adv}^S\right), \#\omega^*\right), \\
where \ S=\left(\alpha,\omega^*\right), \ \omega^*=\mathop{\arg\min} \limits_{\omega} \sum\limits_{(x,y)\in D} \mathcal{L}_{KD}^\alpha(x,y),
\end{gather*}
where $\mathcal{X}$ denotes a searching space.

In this situation, architecture and weight of student model are separately optimized. 
Weight optimization is achieved by performing knowledge distillation on a given architecture, while architecture optimization can be solved by neural architecture search. 
In the proposed framework of \texttt{RoSearch}, we adopt a NAS controller to sample student architectures from search space.
Each searched architecture is distilled with pre-trained language model as teacher model. 
Then its robustness, accuracy and efficiency are evaluated and fed back to controller to guide the search direction.
Algorithm~\ref{alg:rosearch} summarizes the workflow of \texttt{RoSearch}.

Neural architecture search algorithms can be described in three dimensions: Search Strategy, Search Space and Model Evaluation~\cite{survey18}. 
Search space is the feasible range of model architectures, which limits the upper and lower bound of search results.
Search strategy controls how to search in the search space, which determines whether the best architecture can be found in the search space.
Model evaluation method affects the score of each element in the search space, which determines the search direction and target.
A DAG-based search space is built for \texttt{RoSearch} and an evolutionary-algorithm based searching strategy is adopted.
Our robustness-aware model evaluation function $\mathcal {F}\left(\cdot,\cdot,\cdot\right)$ will be described in Section~\ref{subsec:modeleval}.

\begin{algorithm}[t]
\caption{\texttt{RoSearch} Framework}
\label{alg:rosearch}
\KwIn{Teacher model $T$, training dataset $D_{train}$, evaluation dataset $D_{val}$.}
\KwOut{Student model $S^*$.}

Define search space $\mathcal{X}$

\While{$Fitness(S)$ not converge}{
    Controller samples an architecture $\alpha$ from $\mathcal{X}$
    
    \tcc{Perform knowledge distillation}
    $\mathcal{L}^\alpha_{KD}(T, D_{train}) \rightarrow $ update $\omega$ 
    
    $S \leftarrow (\alpha, \omega)$
    
    \tcc{Student model evaluation}
    Generate adversarial dataset $D_{adv}$ from $D_{val}$
    
    Accuracy score $P_{acc} \leftarrow \mathcal{A}_S\left(D_{val}\right) $ 
    
    Robustness score $P_{rob} \leftarrow \mathcal{A}_S\left(D_{adv}\right) $ 
    
    Efficiency score $P_{eff} \leftarrow \#\omega $ 
    
    $Fitness(S) \leftarrow \mathcal{F}\left(P_{acc},P_{rob},P_{eff}\right)$
    
    \If{$Fitness(S)>Fitness\left(S^*\right)$}{
    $S^* \leftarrow S$
    }
    
    Feed $Fitness(S)$ back to controller
    
}

\end{algorithm}

\subsection{Search Strategy}
In \texttt{RoSearch}, an evolutionary algorithm based searching strategy is utilized because of its simplicity and satisfactory performance in~\cite{evo} and~\cite{evo_trans}. 
Moreover, architectures are seem as individuals in evolutionary algorithm, each model is independent during search, which is convenient for architecture analysis. 

Evolutionary algorithm can be mainly divided into four steps: (1) population initialization, (2) fitness evaluation, (3) natural selection, (4) evolution operation.
Population is a set of individuals.
In population initialization step, we select the simplest architecture in search space and generate architectures by making random evolution operations as the initial individuals. 
In fitness evaluation step, each searched model is knowledge distilled and evaluated in order to obtain its fitness to environment.
Tournament Selection Algorithm~\cite{tour} is adopted in natural selection step. 
A certain number of individuals in population are randomly sampled for competition with each other. 
The architecture with highest fitness is selected to generate an offspring in evolution operation step and the one with lowest fitness is eliminated. 
Evolution operation step takes an architecture as input and generates a similar architecture in search space. 
The offspring architecture is then added to population and whole process is repeated from fitness evaluation step. 
Searching continues until fitness converges.

\subsection{Student Architecture and Search Space}

\begin{figure}[t]
\begin{center}
\centerline{\includegraphics[width=\columnwidth]{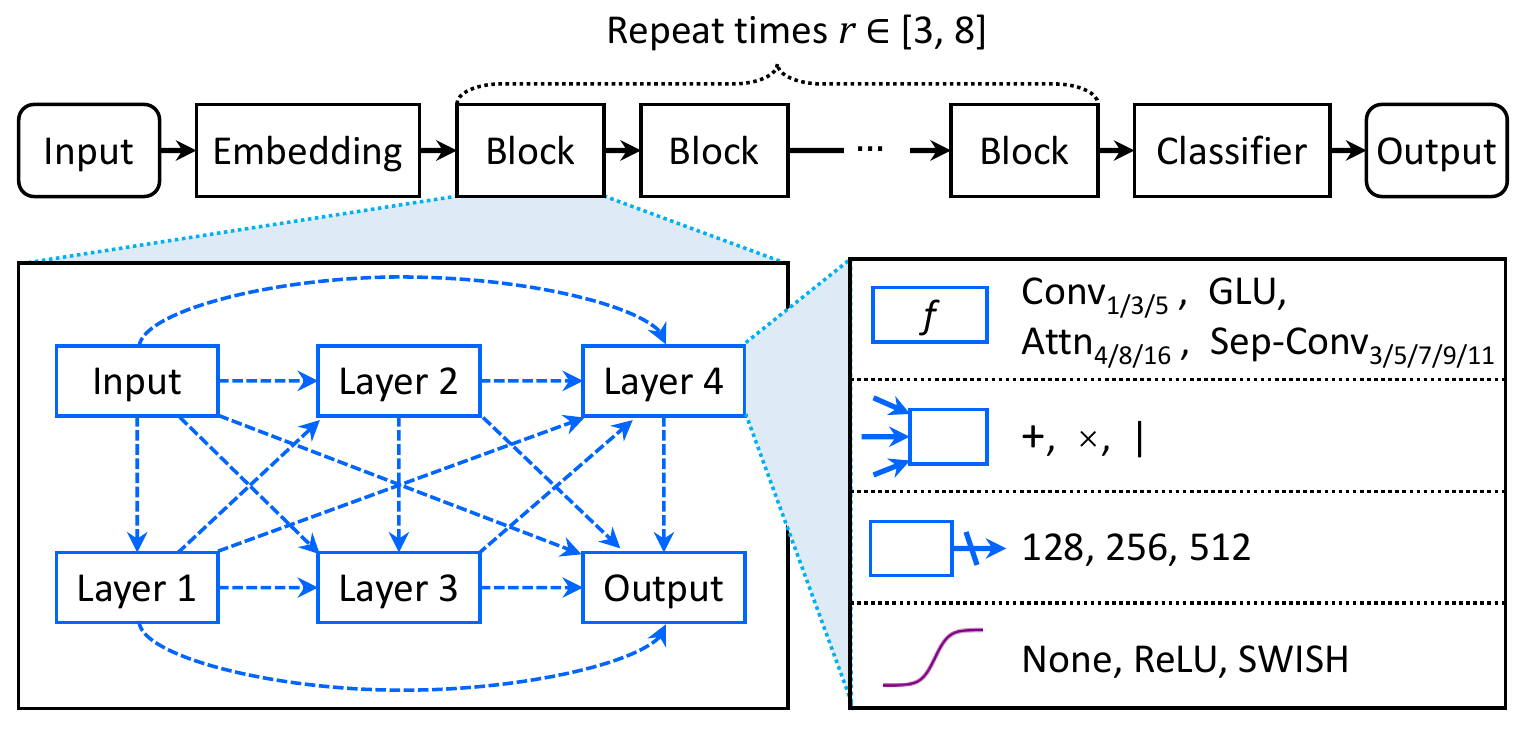}}
\caption{Student architecture demonstration and DAG based search space for repeated Blocks.}
\label{fig:space}
\vspace{-4mm}
\end{center}
\end{figure}

\begin{figure}[t]
\begin{center}
\begin{subfigure}{0.23\textwidth}
  \centering
  \includegraphics[width=\linewidth]{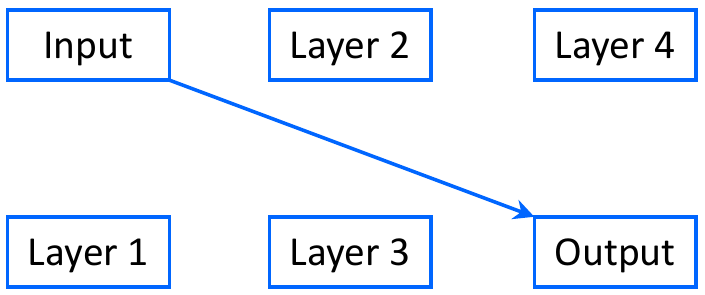}
  \caption{Simplest architecture}
  \label{fig:space_pattern_1}
\end{subfigure}%
\hfill
\begin{subfigure}{0.23\textwidth}
  \centering
  \includegraphics[width=\linewidth]{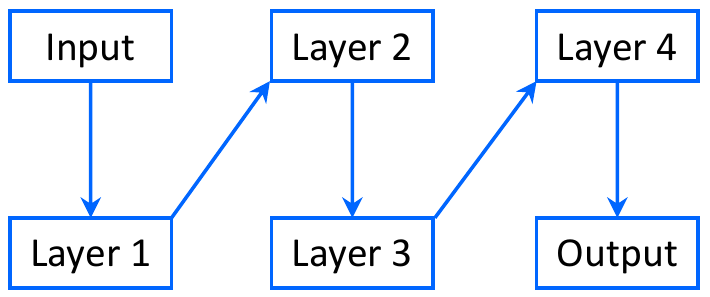}
  \caption{Sequential architecture}
  \label{fig:space_pattern_2}
\end{subfigure}
\begin{subfigure}{0.23\textwidth}
  \centering
  \includegraphics[width=\linewidth]{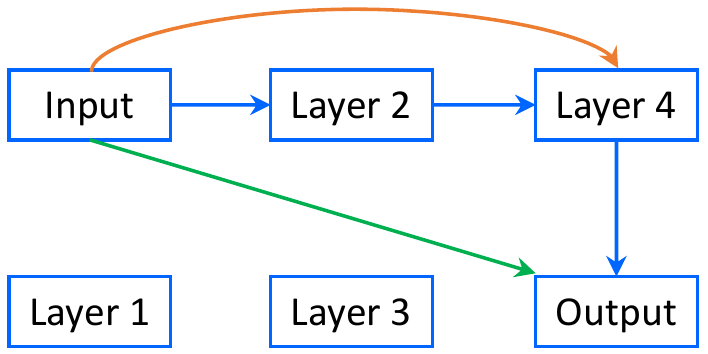}
  \caption{Residual connection between Blocks (green) or Layers (orange)}
  \label{fig:space_pattern_3}
\end{subfigure}%
\hfill
\begin{subfigure}{0.23\textwidth}
  \centering
  \includegraphics[width=\linewidth]{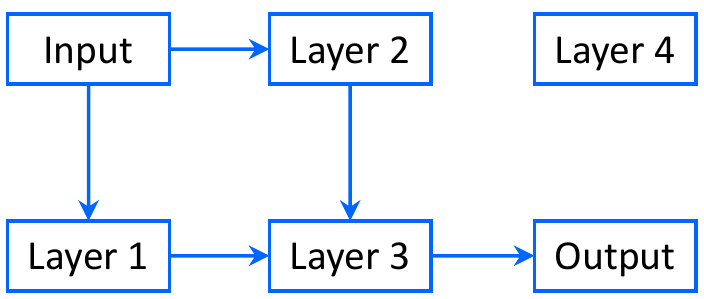}
  \caption{Branch architecture}
  \label{fig:space_pattern_4}
\end{subfigure}
\caption{Representation of some typical architecture patterns in search space.}
\label{fig:pattern}
\vspace{-4mm}
\end{center}
\end{figure}

Since relation between architectures and robustness are not clear yet, we design a search space with high degree of freedom.
Figure~\ref{fig:space} presents the architecture of student model in \texttt{RoSearch}. 
Student model contains Embedding layer, repeated Blocks and Classifier from input to output. 
Architecture of repeated Blocks is represented as directed acyclic graph $G=(V,E)$, where $V=\{v_0,v_1,...,v_{n-1}\}$, $v_0$ represents input node, $v_{n-1}$ represents output node, $v1,...,v_{n-2}$ represents computational layer node, $n$ is a hyper-parameter denoting number of nodes ($n=6$ by default). Edges in DAG represents data-flow, and only edges from vertex with lower ID to higher ID are allowed, i.e. $E \subset \{(v_i,v_j)|i<j\}$. 
As shown in Figure~\ref{fig:pattern}, various typical architecture patterns can be described in this representation, such as residual connections or branch architectures, etc.
Same as BERT, Embedding layer consists of word embedding, position embedding and segment embedding. 
Classifier consists of Max Pooling layer and Fully Connected layer. 
Architecture of these two parts remain unchanged in searching process except for adapting width of hidden states to Blocks. 

Search space can be described as Cartesian Product of value ranges for all variable dimensions in student architecture.
Repeat times of Blocks varies in range \{3, 4, 5, 6, 7, 8\}, and width of hidden states between Blocks varies in range \{128, 256, 512\}, which leads searched architectures lie in a lightweight zone. 
Sub search space of Block can be parameterized by properties of every computational nodes and output node, as well as data-flow connection patterns.
Properties of each computational node consists of Layer Type, Layer Parameter, Output Width, Input Mode and Activation Function, output node only varies in Input Mode and Activation Function.
Search space of data-flow connection in a Block is the power set of all allowed edges in DAG. 
Layer Type of computational nodes ranges in \{\texttt{Conv}, \texttt{Sep-Conv}, \texttt{Attn}, \texttt{GLU}\}.
\texttt{Conv} denotes Convolution Layer, widely used in CNN NLP models.
\texttt{Sep-Conv} denotes Depth Separable Convolution Layer~\cite{mobilenets}, which is a lightweight variant of Convolution Layer.
\texttt{Attn}, to be specific, denotes Self-Attention Layer, which plays an important role in Transformer~\cite{transformer} based models like BERT, and supplies expression ability of word relationship without range limit.
\texttt{GLU} denotes Gated Linear Units Layer~\cite{glu}, which introduces extra nonlinear operation and performs well in~\cite{evo_trans}. 
For \texttt{Conv} layers, Layer Parameter ranges in \{1, 3, 5\}, represents width of convolution kernel. 
When 1 is taken as Layer Parameter, it represents a Linear Layer. 
For \texttt{Sep-Conv} layers, Layer Parameter representing kernel width ranges in \{3, 5, 7, 9, 11\}. 
For \texttt{Attn} layers, Layer Parameter ranges in \{4, 8, 16\}, representing the number of attention head. 
\texttt{GLU} layers have no Layer Parameter. 
Output Width ranges in \{128, 256, 512\}. 
Input Mode ranges in \{\texttt{Add}, \texttt{Mul}, \texttt{Concat}\}, where \texttt{Add} denotes all the inputs of such node will be summed (padded to same width with 0), \texttt{Mul} denotes all the inputs will be multiplied (padded to same width with 1), \texttt{Concat} denotes all the inputs will be concatenated along hidden dimension. 
Activation Function ranges in \{\texttt{None}, \texttt{ReLU}~\cite{relu}, \texttt{SWISH}~\cite{swish}\}.

When search space determined, evolution operation in search strategy can be defined. 
Evolution Operation takes an architecture in search space as input, randomly generates a new architecture similar to it and guarantees the new architecture is in the search space. 
Formally, evolution operation $Evo(\cdot)$ satisfies:
\begin{gather*}
\alpha_1 \in \mathcal{X} \ \Rightarrow \  \big( \left(Evo\left(\alpha_1\right) \in \mathcal{X} \right) \  and \  \left(  0 < d \left( Evo\left(\alpha_1\right), \alpha_1\right) < \epsilon \right) \big), 
\end{gather*}
where $d(\cdot,\cdot)$ denotes the distance between two architectures, $\epsilon$ denotes a specified distance tolerance.

In our design, Evolution Operations include: (1) randomly change in repeat times of Blocks, (2) randomly change in width of hidden states between Blocks, (3) randomly add or remove an edge from DAG, (4) randomly change in Layer Type, Layer Parameter, Output Width, Input Mode or Activation Function of a random node. 
Since some Evolution Operations may cause conflict when building a neural network, extra mutations should be considered. 
When Layer Type changes, Layer Parameter of same node is randomly selected from the parameter range of new Layer Type.
Adding edge is only allowed when number of edges is less than $\left(\frac{n(n-1)}{2}-3\right)$.
Removing edge is only allowed when number of edges is greater than $3$. 
After adding or removing edges, connectivity of DAG is checked. 
Nodes with only input/output edges will be connected to a random output/input edge. 

\subsection{Robustness-aware Model Evaluation}
\label{subsec:modeleval}

In \texttt{RoSearch}, we propose a robustness-aware model evaluation metric to grade each model $S$ in search process. 

Each student architecture sampled from search space is constructed as a neural network and then performed knowledge distillation first.
To overcome the obstacle between layer-wise knowledge distillation and variable architecture in searching, Probe Network technique~\cite{adabert} is adopted to calculate layer-wise knowledge transfer loss when performing distillation.
Classifiers are trained for each layer's output in teacher model as teacher probe networks. 
Student probe networks are trainable classifiers for output of each Block in student model. 
Cross Entropy Losses between predictions of each student probe network and corresponding teacher probe network are summed as knowledge transfer loss.
In addition, MSE loss between teacher model's and student model's logits, as well as traditional knowledge distillation loss are also adopted.

After distillation, each model is evaluated in three dimensions: Accuracy, Robustness and Efficiency.

\textbf{Accuracy}.
Accuracy is simply measured by model's correct rate evaluated on benign datasets.

\textbf{Robustness}.
To evaluate the robustness of distilled student models, TextFooler~\cite{textfooler} is used to generate adversarial examples for each student model. 
Accuracy on adversarial examples of searched models is calculated as robustness metric. 
Since the robustness under transfer attack significantly differs from the robustness under direct attack and no correlation between them is observed, direct attack on each model is performed independently during searching to give a more precise evaluation.

\textbf{Efficiency}.
In order to guide searching process towards lightweight models, an efficiency aware metric is included in the model evaluation function.
Model's efficiency is measured by the amount of parameters.
Less parameters indicates higher efficiency.

Three metrics for three attributes are weighted summed as the total score for guiding the searching process, so that trade-off between accuracy, robustness and efficiency could be performed to fit different scenarios. 
The whole fitness evaluation function is summarized as:
\begin{gather*}
\label{eq:fitness}
Fitness(S)=\mu_1\mathcal{A}_S(D)+\mu_2\mathcal{A}_S\left(TextFooler(D)\right)+\mu_3\left(-\#\omega_S\right),
\end{gather*}
where $S$ represents the student model to be evaluated, $D$ represents the involved dataset, $\#\omega_S$ represents the number of parameters in model $S$.

\section{Experiments}
\label{sec:experiments}

In this section, the proposed \texttt{RoSearch} is evaluated on natural language classification datasets and compared with existing baseline methods.
Further experimental results on SST-2 dataset is presented for robustness analysis. 
In addition, we summarize the relationship between robustness of student models and model architecture properties through statistics of searched models.

\begin{table}[t]
\caption{
Result comparison on accuracy, robustness and efficiency of our approach and existing baselines. Model with (AT) means that it is obtained by adversarial training in fine-tuning stage.}
\label{tab:res_dataset}
\vspace{-2mm}
\begin{center}
\begin{small}
\begin{sc}
\begin{tabular}{|l|l|c|c|c|}
\hline
Dataset	& Model & $\mathcal{A}$	& $\mathcal{A}\left(D_{adv}\right)$ & $\#\omega$\\
\hline
\hline
\multirow{7}{*}{SST-2}
 & BERT	& 92.3\%	& 12.4\%	& 110M\\ 
 & BERT(AT)	& 92.1\%	& 21.3\%	& 110M\\ \cline{2-5} 
 & PKD	& 90.7\%	& 6.2\%	& 67M\\ 
 & PKD(AT)	& 83.7\%	& 14.0\%	& 67M\\ \cline{2-5} 
 & AdaBERT-SST2	& 89.0\%	& 7.1\%	& 6.4M\\
 & AdaBERT-SST2(AT)	& 73.4\%	& 22.0\%	& 6.4M\\ \cline{2-5} 
 & Ours	& 84.1\%	& 45.8\%	& 24M\\
\hline
\hline
\multirow{5}{*}{Ag News}
 & BERT	& 95.0\%	& 27.0\%	& 110M\\ 
 & BERT(AT)	& 94.0\%	& 35.8\%	& 110M\\ \cline{2-5} 
 & PKD	& 94.8\%	& 18.7\%	& 67M\\
 & PKD(AT)	& 90.0\%	& 19.3\%	& 67M\\ \cline{2-5} 
 & Ours	& 90.3\%	& 47.8\%	& 16.9M\\
\hline
\end{tabular}
\end{sc}
\end{small}
\end{center}
\end{table}

\subsection{Setup}

\textbf{Datasets.} 
SST-2~\cite{sst2} and Ag News~\cite{agnews} datasets are adopted for evaluation in this work.
SST-2 dataset of GLUE~\cite{glue} benchmark aims to classify movie reviews into positive or negative categories according to their sentiment.
Ag News dataset is a four-category classification task, which contains tens of thousands pieces of news. 
Each news is categorized into one of the four topics: World, Sports, Business and Sci/Tech. 
For each dataset, training set is utilized to fine-tune teacher model and obtain student model with knowledge distillation. 
Testing set is adopted to evaluate accuracy and robustness for student models. 

\textbf{Implementation Details.} 
For the evolutionary search strategy in \texttt{RoSearch}, population size is set to $100$. 
In each generation, $2$ individuals are randomly selected for competition in tournament selection algorithm. 
Weighting factors in fitness equation were set to $\mu_1=1$, $\mu_2=1$, $\mu_3=2$. 
Accuracy and robustness in the fitness function are measured in percentage.
Number of weight parameters is plugged in millions (M). 
Similar score threshold of TextFooler is set to $0.7$ for robustness evaluation.
For the knowledge distillation process of searched models, we leverage fine-tuned $\rm BERT_{BASE}$ with probe networks as teacher model. 
Vanilla knowledge distillation loss, probe network distillation loss, MSE loss between logits of student and teacher are combined with weighting factors.
The three weighting factor are set as $0.5$, $0.25$ and $0.25$ respectively. 
During distillation, temperature of vanilla distillation loss is set to $2$, learning rate is set to $5e-4$, epoch number is set to $15$.
Prediction of teacher model required for loss calculation is only related to input data instead of each student model. 
Thus, teacher model fine-tuning, probe networks training, and prediction on input data are performed before searching process. 

Our algorithm are implemented with PyTorch \cite{pytorch} framework and deployed on a Slurm \cite{slurm} task management platform. About sixteen Nvidia V100 GPUs are utilized as workers. 

\begin{figure}[t]
\begin{center}
\centerline{\includegraphics[width=\columnwidth]{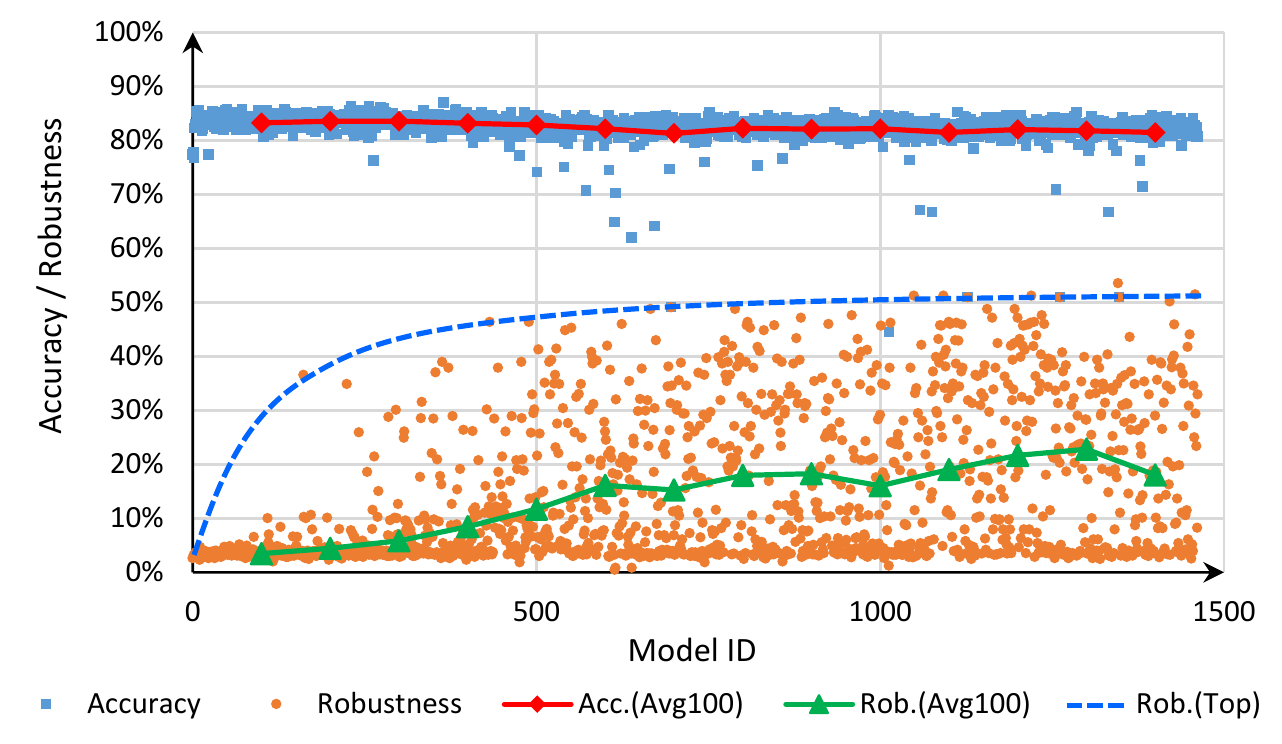}}
\caption{Searching process on SST-2 dataset. Accuracy of searched models remains while average and upper-bound of robustness gradually increase.}
\label{fig:result}
\end{center}
\end{figure}

\subsection{Comparison Results}

\textbf{Baselines.} 
Four categories of models are adopted as baselines for comparison: (1) pre-trained language model without adversarial training, i.e. the teacher model $\rm BERT_{BASE}$~\cite{bert}, (2) pre-trained language model with adversarial training in fine-tuning stage, i.e. BERT(AT), (3) student model derived from knowledge distillation without adversarial fine-tuning, i.e. PKD~\cite{pkd} and AdaBERT-SST2~\cite{adabert} (only for SST-2 dataset), (4) student model derived from knowledge distillation with adversarial fine-tuning, i.e. PKD(AT) and AdaBERT-SST2(AT).

These models are compared in three dimensions: accuracy, robustness and efficiency.
As shown in Table~\ref{tab:res_dataset}, our approach obtains significant robustness improvement with low accuracy drop, compared with the teacher model and existing distillation methods, even if adversarial training tricks are adopted.
Furthermore, our compression ratio of model size could reach up to $4.6\times$\textasciitilde$6.5\times$ than teacher model $\rm BERT_{BASE}$.
Though high computation cost is needed for our method, it remains in training stage.

Further evaluation details of \texttt{RoSearch} on SST-2 dataset are illustrated in Figure~\ref{fig:result}.
During the searching process, student model's accuracy is located at $80\%$\textasciitilde$85\%$.
For robustness, both average value and trend of top robust models are gradually increasing along searching process. The diagram proves the effectiveness of our method on improving robustness of student models while maintaining the accuracy.

\begin{figure}[t]
\begin{center}
\begin{subfigure}{0.23\textwidth}
  \centering
  \includegraphics[width=\linewidth]{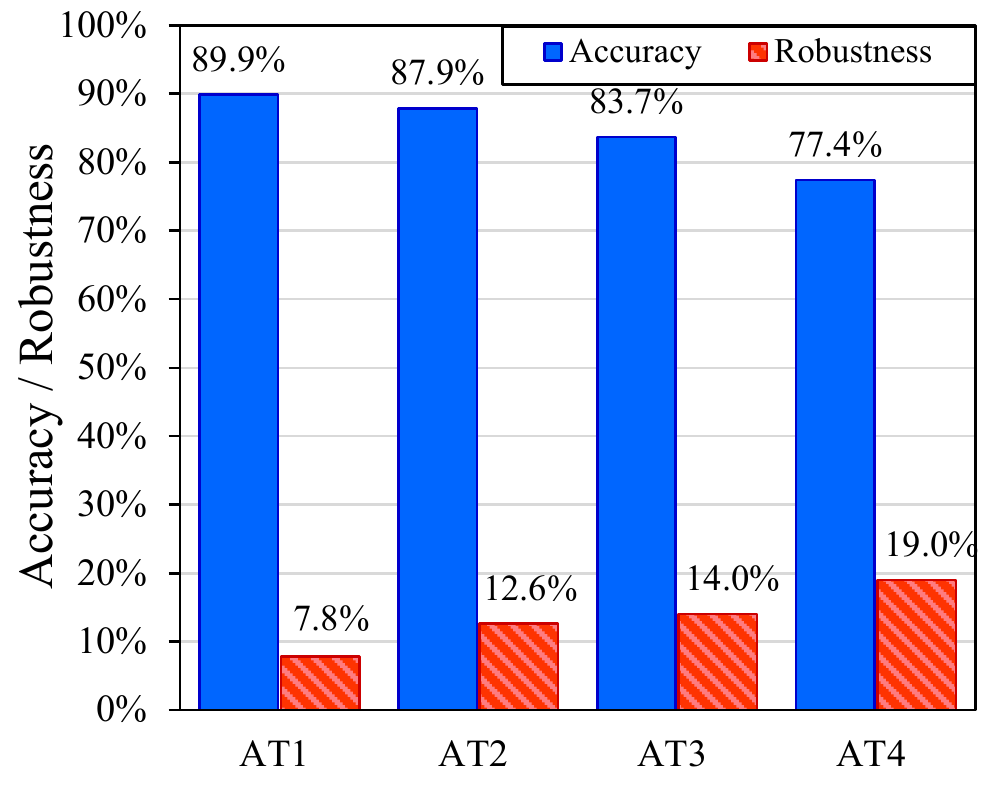}
  \caption{Adversarial training of PKD with different strength}
  \label{fig:at_ours_1}
\end{subfigure}%
\hfill
\begin{subfigure}{0.23\textwidth}
  \centering
  \includegraphics[width=\linewidth]{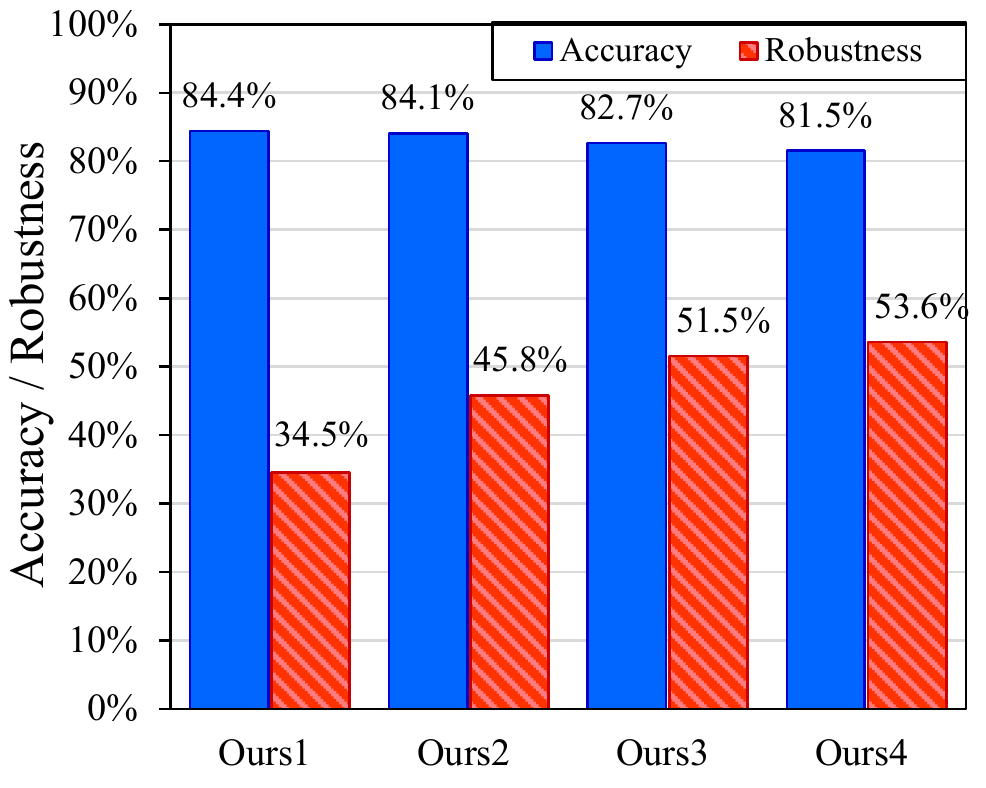}
  \caption{Searched models of our proposed \texttt{RoSearch}}
  \label{fig:at_ours_2}
\end{subfigure}
\caption{Models' accuracy and robustness comparison between adversarial training and \texttt{RoSearch}.}
\label{fig:at_ours}
\vspace{-4mm}
\end{center}
\end{figure}

Adjusting the strength of adversarial training will affect the trade-off between model robustness and accuracy. On the other hand, student models with similar scores in our searching process may also differ in the robustness and accuracy terms. We draw robustness-accuracy curve of these methods in Figure~\ref{fig:at_ours} to give a more detailed comparison.

\subsection{Robust Architecture Analysis}

\begin{figure}[t]
\begin{center}
\centerline{\includegraphics[width=\columnwidth]{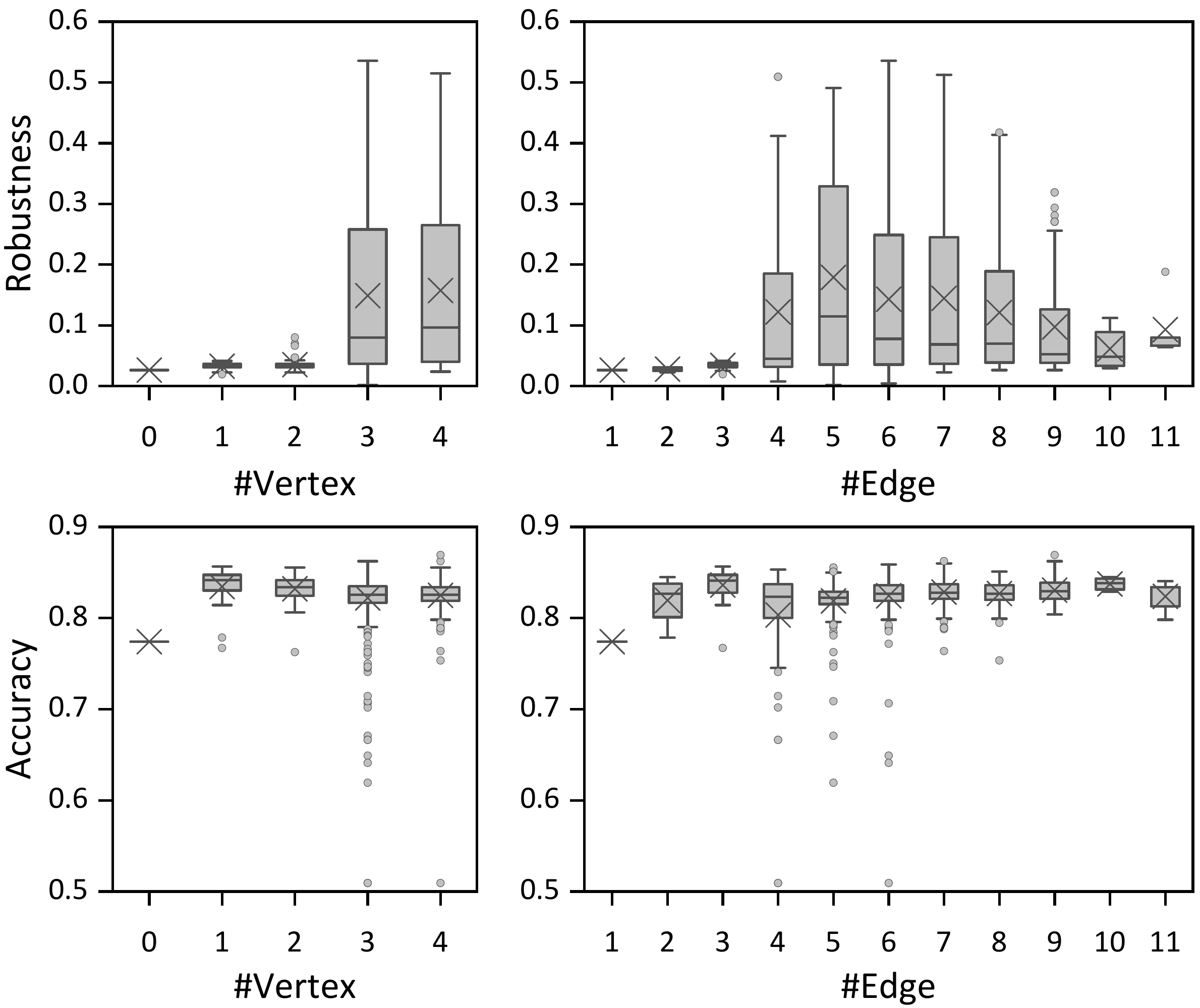}}
\caption{Statistics on robustness and accuracy of student architectures according to DAG properties. \#Vertex denotes number of computational nodes that are connected in DAG. \#Edge denotes number of activated edges in DAG.}
\label{fig:v_e}
\vspace{-5mm}
\end{center}
\end{figure}

According to experiment on SST-2 dataset, we statistically summarize the relationship between architectures and the robustness of student models, in order to give an answer to Question {\large \textcircled{\small \textbf{3}}} in Section~\ref{sec:intro}. 

Figure~\ref{fig:v_e} shows the influence of DAG properties on robustness and accuracy of student model. The number of vertices and edges have no obvious effect on the accuracy except that least edges and vertices cause lowest average accuracy. On the other hand, student architectures with more than 3 vertices in DAG have a significant improvement on robustness. Robustness increases when number of active edges in DAG come up to 4, and begins to decrease when number of edges is greater than 8.
It can be summarized that student architectures with higher layer number and moderate connectivity tend to have a higher robustness.

\begin{figure}[t]
\begin{center}
\centerline{\includegraphics[width=\columnwidth]{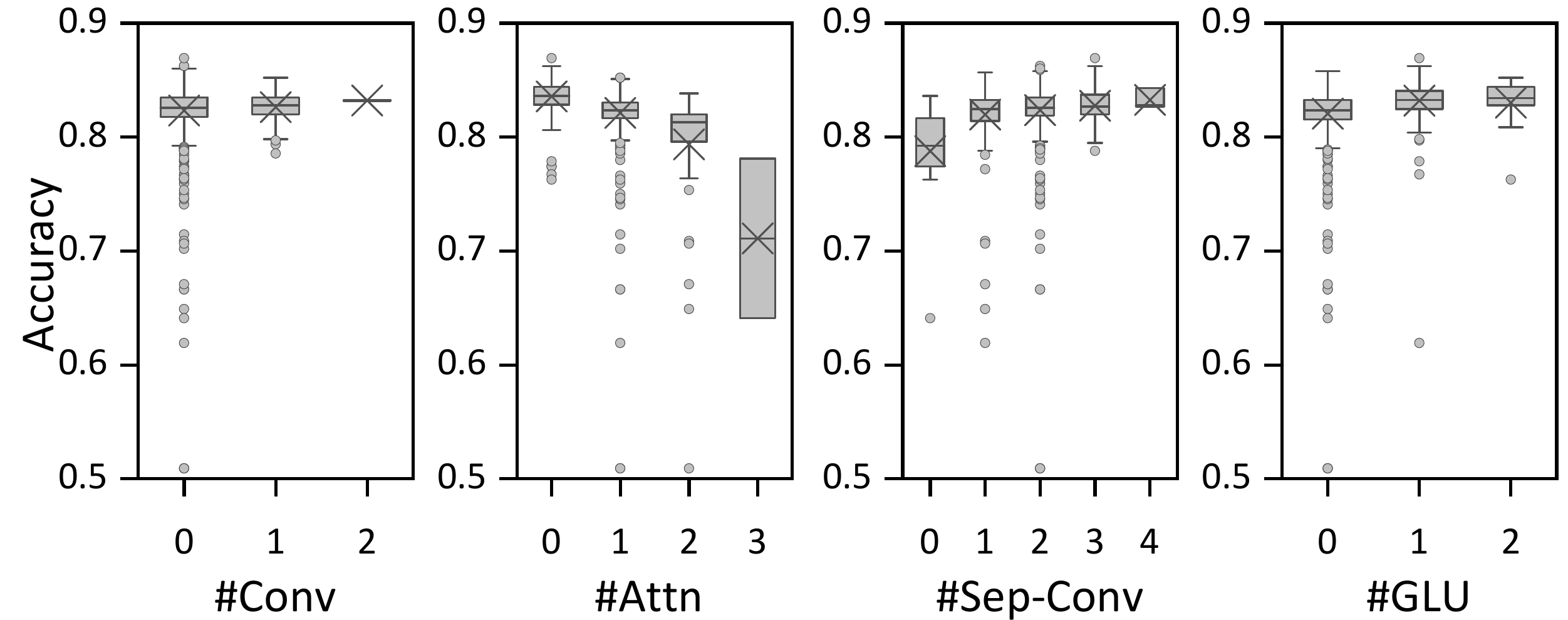}}
\caption{Relationship between accuracy and number of different types of layers used in a Block.}
\label{fig:acc_type}
\vspace{-5mm}
\end{center}
\end{figure}

\begin{figure}[t]
\begin{center}
\centerline{\includegraphics[width=\columnwidth]{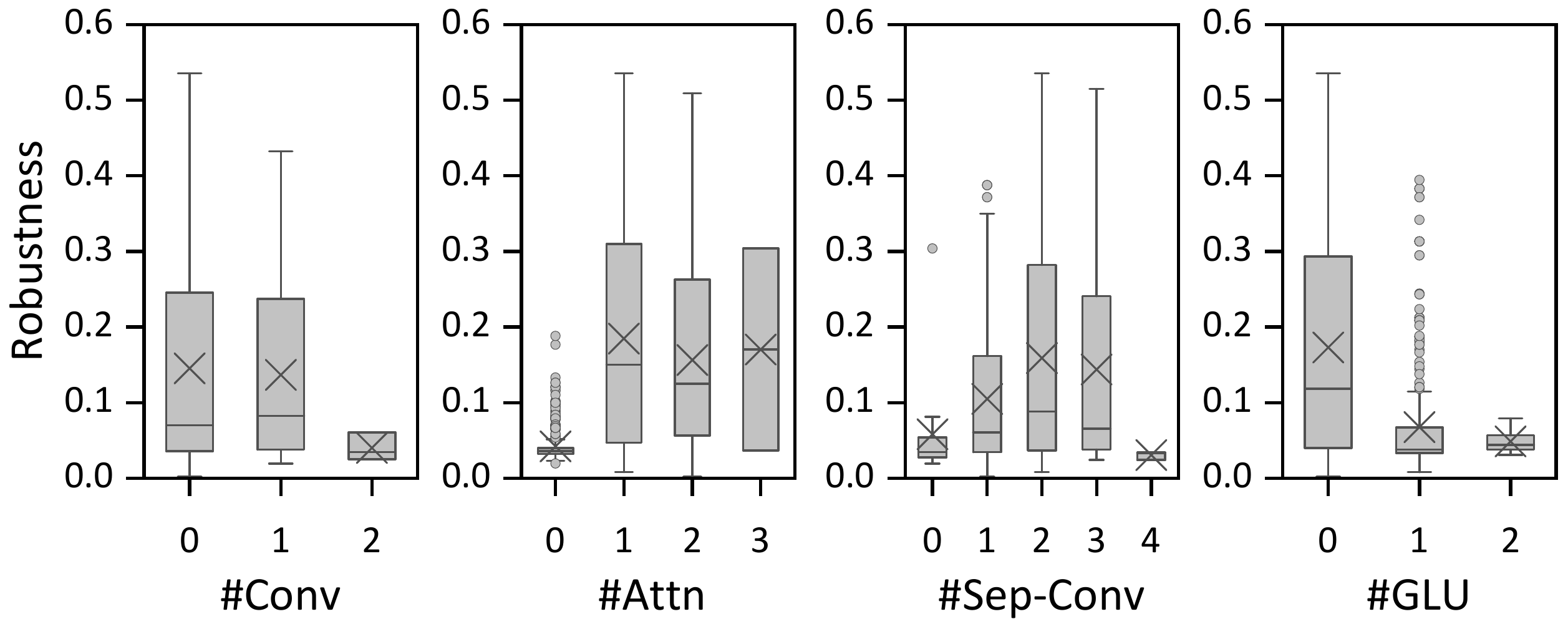}}
\caption{Relationship between robustness and number of different types of layers used in a Block.}
\label{fig:rob_type}
\vspace{-5mm}
\end{center}
\end{figure}

We also analyzed the influence of layer with different type on accuracy and robustness, results are summarized in Figure~\ref{fig:acc_type} and Figure~\ref{fig:rob_type} respectively. From Figure~\ref{fig:acc_type}, it can be concluded that there is a positive correlation between number of \texttt{Conv}, \texttt{Sep-Conv} and \texttt{GLU} layers and accuracy of student models, while number of Attention layers is negatively correlated to the accuracy. For robustness, number of \texttt{Conv} and \texttt{GLU} layers provide negative contribution. Number of Attention layers and \texttt{Sep-Conv} layers give positive contribution when number is moderate, however, excessive number of same type reduces the positive influence on robustness.

\section{Conclusions}
\label{sec:conclusions}

In this work, we point out the security problem of existing knowledge distillation methods for pre-trained language models. To overcome the problem, we focus on an architecture view of student models.
We propose an algorithm \texttt{RoSearch} based on evolutionary NAS strategy to search for robust student architecture, while maintaining the accuracy of pre-trained language models and efficiency acquired from knowledge distillation. 
Each student architecture sampled from a DAG based search space is distilled and evaluated in robustness, accuracy and efficiency metrics. 
Experimental results show the effectiveness of our algorithm. We also summarize the relationship between student architecture and the robustness based on statistics of searched models.



\bibliographystyle{unsrt}
\bibliography{rosearch}



\end{document}